\documentclass{article}
\usepackage{spconf,amsmath,graphicx,caption,subcaption,tabularx,lipsum}
\usepackage{hyperref}
\hypersetup{colorlinks,allcolors=black}
\usepackage{amssymb}
\usepackage[cal=zapfc]{mathalfa}
\usepackage{subcaption}

\title{Lattention: Lattice-attention in ASR rescoring}
\name{Prabhat Pandey\sthanks{Equal Contribution}, Sergio Duarte Torres\footnotemark[1], Ali Orkan Bayer, Ankur Gandhe, Volker Leutnant}
\address{Amazon Alexa AI}

\begin{document}
\ninept

\maketitle

\let\thefootnote\relax\footnotetext{© 2021 IEEE. Personal use of this material is permitted. Permission from IEEE must be obtained for all other uses, in any current or future media, including reprinting/republishing this material for advertising or promotional purposes, creating new collective works, for resale or redistribution to servers or lists, or reuse of any copyrighted component of this work in other works.}

\begin{abstract}
  Lattices form a compact representation of multiple hypotheses generated from an automatic speech recognition system and have been shown to improve performance of downstream tasks like spoken language understanding and speech translation, compared to using one-best hypothesis. In this work, we look into the effectiveness of lattice cues for rescoring n-best lists in second-pass. We encode lattices with a recurrent network and train an attention encoder-decoder model for n-best rescoring. The rescoring model with attention to lattices achieves 4-5\% relative word error rate reduction over first-pass and 6-8\% with attention to both lattices and acoustic features. We show that rescoring models with attention to lattices outperform models with attention to n-best hypotheses. We also study different ways to incorporate lattice weights in the lattice encoder and demonstrate their importance for n-best rescoring.
\end{abstract}

\section{Introduction}
\label{sec:intro}

In a typical multi-pass automatic speech recognition (ASR) system, the first-pass system produces lattices \cite{ney1994word} or n-best hypotheses \cite{schwartz1991comparison} which are rescored in the second-pass. More commonly, a neural language model (NLM) trained on large amount of text data is used in the second-pass rescoring \cite{mikolov2010recurrent, sundermeyer2012lstm}. Recently, stronger rescoring models utilizing acoustic information have been proposed. In \cite{sainath2019two}, a listen-attend-spell \cite{chan2016listen} based model was proposed to rescore n-best lists where the encoder is shared with the first-pass recurrent neural network transducer (RNN-T) \cite{graves2012sequence} model. Similarly, in \cite{gandhe2020audio}, NLM was extended to attend to audio features generated by the acoustic model in the first-pass ASR system. In further extension to \cite{sainath2019two}, a deliberation network \cite{xia2017deliberation} based model with additional attention to n-best hypotheses was introduced in \cite{hu2020deliberation}. A more compact representation of the first-pass decoding output are lattices. Lattices encode multiple hypotheses in a condensed form and carry the uncertainties from the first-pass decoding. Using a lattice encoder instead of the 1-best output has been shown to improve performance of downstream tasks like speech translation \cite{sperber2017neural,su2017lattice,sperber2019self,xiao2019lattice} and spoken language understanding \cite{ladhak2016latticernn,zhang2018chinese}.

There has been some previous work on rescoring lattices in second-pass \cite{liu2014efficient,xu2018pruned} instead of a subset of hypotheses in the n-best list, making use of the richer information in the lattices. In this work, we utilize lattice information for n-best rescoring by encoding them with a recurrent network. We train an attention based encoder-decoder model which attends to the lattice encoder and run the decoder in the teacher-forcing mode to rescore n-best lists. We experiment with different encoders: 1-best, n-best, lattice and audio features extracted from the first-pass model. We employ minimum word error rate (MWER) \cite{hori2016minimum} training criterion which has been shown to improve accuracy of attention-based rescoring models \cite{sainath2019two,gandhe2020audio,hu2020deliberation}.

There has already been some work on representing lattice structures in recurrent encoders \cite{sperber2017neural,su2017lattice,ladhak2016latticernn} and transformers \cite{sperber2019self,xiao2019lattice} models. In \cite{sperber2017neural}, LatticeLSTM was proposed for machine translation, which extends TreeLSTM \cite{tai2015improved} to encode directed acyclic graphs with weights. We utilize LatticeLSTM with certain modifications as the lattice encoder for ASR n-best rescoring in this work. Specifically, following are the contributions of this paper: (1) We propose a simplified method for encoding lattice weights with similar performance as \cite{sperber2017neural}, (2) We show that lattice-attention rescoring model can provide 4-5\% relative word error rate reduction (WERR) over first-pass, (3) LatticeLSTM-based lattice encoder results in more improvements compared to n-best deliberation encoder \cite{hu2020deliberation}, even for lattices containing same hypotheses as the n-best, (4) Attending to both audio and lattice further reduces word error rate (WER), resulting in 6-8\% relative WERR over first-pass, (5) We study the effect of different mechanisms to incorporate lattice weights in LatticeLSTM and show that unweighted lattice encoders (TreeLSTM) are detrimental for attention-based models and integrating lattice weights is important to mitigate confusions arising from contradictory lattice arcs.

\section{Lattice-attention Model}
\label{sec:lattice_encoder}

\subsection{Lattice representation}
\vskip -0.1cm

\begin{figure}[t]
  \centering
  \includegraphics[scale=0.37]{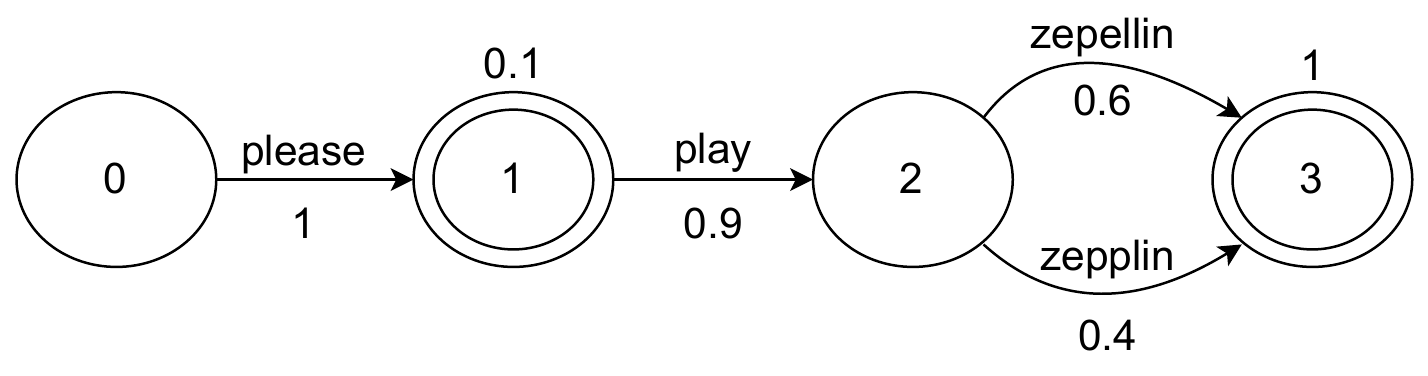}
  \caption{Example of WFST representation of a edge-labeled-lattice with forward-normalized weights.}
  \label{fig:lattice_compact}
  \vskip -0.3cm
\end{figure}

We represent the lattices generated from first-pass ASR system as a weighted finite state transducer (WFST) and apply epsilon removal, determinization, minimization, weight pushing to initial states, removal of total weights and eventually, topological sorting of nodes \cite{allauzen2007openfst}.
The original lattices have labels on its arcs (we refer them as edge-labeled lattice, an example is shown in Figure~\ref{fig:lattice_compact}). In \cite{sperber2017neural}, instead of edge-labeled lattices, node-labeled lattices are used in LatticeLSTM because of its intuitive appeal as hidden states in LatticeLSTM represent a single token in case of node-labeled lattice. We also use node-labeled lattices in this work which are generated by applying line-graph algorithm \cite{hemminger1983line} on edge-labeled lattices. Figure~\ref{fig:lattice_node_based} shows the node-labeled lattice for the edge-labeled lattice depicted in Figure \ref{fig:lattice_compact}. We explain how the weights from edge-labeled-lattices are transformed to node-labeled lattices in Section \ref{section:node_labeled_lattices}.

\subsection{Forward-weight normalization in edge-labeled lattices}
\vskip -0.1cm
\label{section:score_normalization_in_lattices}

The costs on the lattices are usually not in the probability space. So, before converting to node-labeled lattice, we forward-normalize costs in in the edge-labeled lattice so that weights on all arcs outgoing from a node sum to 1. Let $e$ be an edge in the edge-labeled lattice (and a node in the node-labeled lattice), $o(e)$ and $d(e)$ denote the origin and destination nodes of the edge $e$ in the edge-labeled lattice, respectively. Let $O(j)$ denote the edge labels on the outgoing arcs from node $j$ including a dummy edge for final state if $j$ is a final state. We define forward-normalized weight $w_e^\mathcal{F}$ on the edge $e$ as $ w_e^\mathcal{F} = \frac{\sigma(-cost_{e})}{\sum_{j\in O(o(e))} \sigma(-cost_{j})} $, where $\sigma$ is the sigmoid function, $cost_e$ is the first-pass ASR cost on the lattice arc $e$. We represent the dummy edge from a node $j$ which is a final state as $\mathbb{F}(j)$.

\subsection{Weights estimation in node-labeled lattices}
\vskip -0.1cm
\label{section:node_labeled_lattices}

We now define two types of weights for node-labeled lattices which are used in LatticeLSTM. First is \textit{marginal weights} which represent the probability of reaching a node given all the paths that contain that node. Let $I(j)$ denote the set of labels on the incoming arcs to node $j$ in the edge-labeled lattice. Then, marginal weight for a node $e$ in the node-labeled lattice is computed as, $ w_e^\mathcal{M} = \sum_{k\in I(o(e))}w_k^\mathcal{M}\cdot w_e^\mathcal{F}$, using forward-backward algorithm \cite{rabiner1989tutorial}, where $w_e^\mathcal{F}$ is the forward-normalized weight for the corresponding edge $e$ in the edge-labeled lattice. We add a single source node, \textless$s$\textgreater{}, with outgoing arcs to labels on the outgoing arcs of start states in the edge-labeled lattice, and a single sink node, \textless$/s$\textgreater{}, corresponding to the final state in the edge-labeled lattice. Both, $w_{<s>}^\mathcal{M}$ and $w_{</s>}^\mathcal{M}$ evaluate to 1.

\begin{figure}[t]
  \centering
  \includegraphics[scale=0.37]{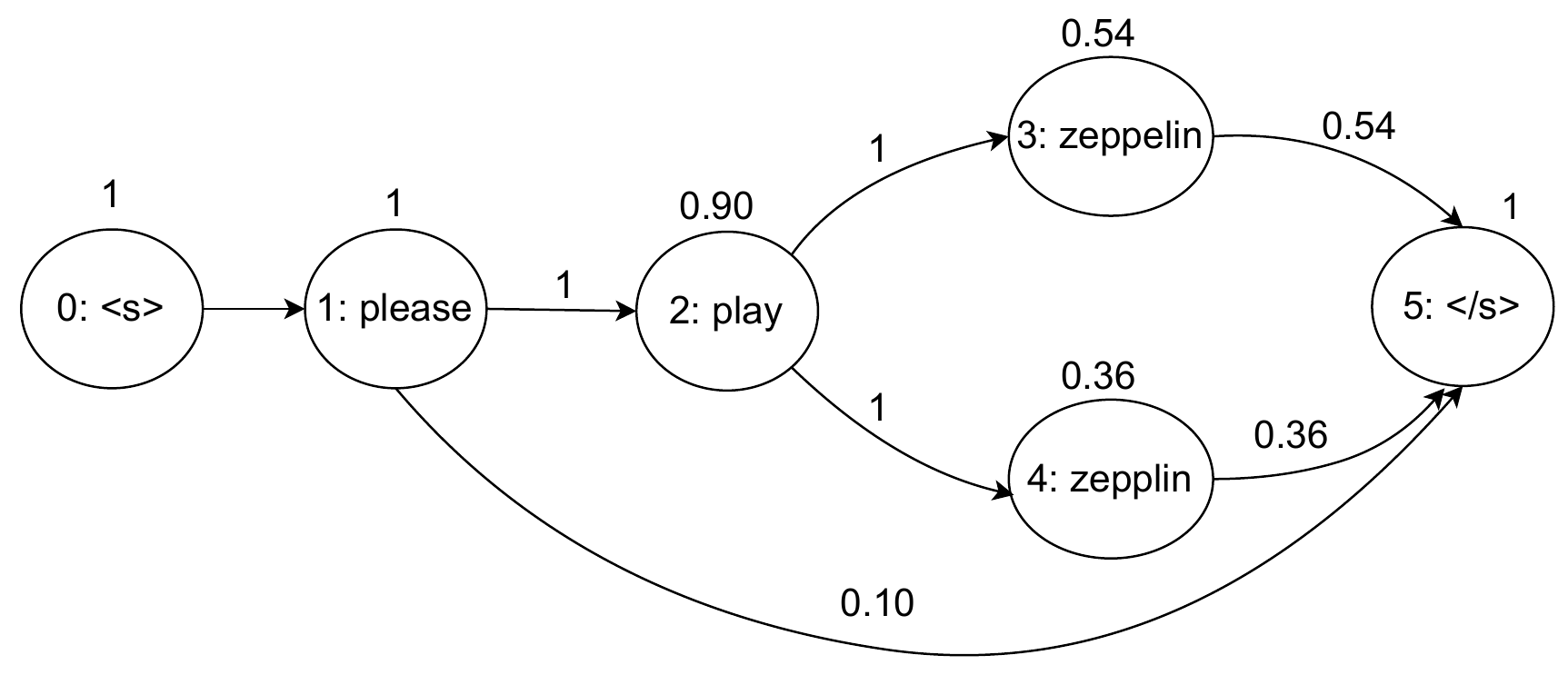}
  \caption{Example of a node-labeled lattice. Marginal weights of lattice nodes are shown on top of each node and backward-normalized incoming arc weights on top of each arc.}
  \label{fig:lattice_node_based}
  \vskip -0.4cm
\end{figure}

To get the relative importance of different incoming arcs to a node, we use \textit{backward-normalized weights},~which are normalized marginal weights for the source node of the arc. Backward-normalized weights for all incoming arcs to a node sum to 1. Formally, for edges $k$ and $e$ in the edge-labeled lattice, if there is an edge from $k$ to $e, e \neq$ \textless$/s$\textgreater{}, in the node-labeled lattice, the backward-normalized weight on the arc from $k$ to $e$ is computed as, $ w_{k,e}^\mathcal{B} = \frac{w_k^\mathcal{M}}{\sum_{j\in I(o(e))}w_j^\mathcal{M}}$. If the destination node of the edge $k$ is a final state in the edge-labeled lattice, the backward-normalized weight on the arc from $k$ to \textless$/s$\textgreater{} in the node-labeled lattice is defined as $w_{k,</s>}^\mathcal{B} = w_k^\mathcal{M}\cdot w_{\mathbb{F}(d(k))}^{\mathcal{F}}$ Figure~\ref{fig:lattice_node_based} shows marginal weights for each node and backward-normalized weights for each arc of a node-labeled lattice.

\subsection{LatticeLSTM}
\vskip -0.1cm
\label{ssec:latticelstm}

\subsubsection{Unweighted LatticeLSTM}
\vskip -0.1cm
\label{ssec:tree_lstm}

Conventional LSTMs use a linear chain network where the hidden state of the network is propagated from the previous node to the next. Although the gates help LSTM to capture long-range dependencies, the linear chain is not well suited for representing linguistic dependencies, which can be better represented by using a tree structured network. TreeLSTMs~\cite{tai2015improved} are designed considering this nature of languages, so that the information is propagated from child nodes to parent nodes. In this paper, we focus on child-sum variant of TreeLSTM that is defined by the following equations, as given in~\cite{tai2015improved}. Given that $P(e)$ is the set of predecessor nodes for node $e$, the memory cell state ($\pmb{c}_e$) and the hidden state ($\pmb{h}_e$) corresponding to node $e$ in the node-labeled lattice are computed as follows, where $W$ and $U$ are weight matrices and $b$ is the bias vector:
\begin{eqnarray}
\tilde{\pmb{h}_e} &=& \sum_{k\in P(e)} \pmb{h}_k
\label{eqn:hsum} \\
\pmb{i}_e&=&\sigma (W^{i}\pmb{x}_e + U^i \tilde{\pmb{h}_e} + \pmb{b}^i) \\
\pmb{f}_{k,e}&=&\sigma (W^f\pmb{x}_e + U^f\pmb{h}_k + \pmb{b}^f) \label{eqn:fgate} \\
\pmb{o}_e&=&\sigma (W^o\pmb{x}_e + U^o\tilde{\pmb{h}_e} + \pmb{b}^o) \\
\pmb{u}_e&=&tanh(W^u\pmb{x}_e + U^u\tilde{\pmb{h}_e} + \pmb{b}^u) \\
\pmb{c}_e &=& \pmb{i}_e \odot \pmb{u}_e + \sum_{k\in P(e)} \pmb{f}_{k,e} \odot \pmb{c}_k \\
\pmb{h}_e &=& \pmb{o}_e \odot tanh(\pmb{c}_e)
\end{eqnarray}

The TreeLSTM cell captures dependencies of incoming arcs by summing uniformly over hidden states of all predecessor nodes. The lattices generated from ASR decoding have scores on its arcs which provide information about the likelihood of different paths. In \cite{sperber2017neural}, TreeLSTM was extended to define LatticeLSTM by incorporating marginal and backward-normalized weights. We explain in detail in the next section.

\subsubsection{Incorporating lattice weights}
\label{ssec:weighted_lattice_lstm}

Similar to \cite{sperber2017neural}, we employ two mechanisms to incorporate backward-normalized weights in the LatticeLSTM cell structure. In the first approach, referred as \textit{weighted child-sum (WCS)}, the uniform-sum of TreeLSTM in Equation~\ref{eqn:hsum} is modified to account for weights on the arc from predecessor nodes:

\begin{equation}
  \tilde{\pmb{h}_e} = \sum_{k\in P(e)} w_{k,e}^\mathcal{B}\cdot \pmb{h}_k
\label{eqn:wcs}
\end{equation}

In the second approach, referred as \textit{biased forget gate (BFG)}, the backward-normalized weights are used to decrease the likelihood, for the cells states corresponding to the predecessor nodes with higher weights, of being attenuated in the forget gate. The forget gate computation in Equation~\ref{eqn:fgate} is modified as follows in LatticeLSTM:

\begin{equation}
  \pmb{f}_{k,e}=\sigma (W^f\pmb{x}_e + U^f\pmb{h}_k + \ln{w_{k,e}^\mathcal{B}} + \pmb{b}^f)
\label{eqn:wfg}
\end{equation}
Additional learnable coefficients, $\pmb{S}_h$ and $\pmb{S}_f$ of same dimension as hidden states, were introduced in \cite{sperber2017neural} and $w_{k,e}^\mathcal{B}$ was transformed to $(w_{k,e}^\mathcal{B})^{\pmb{S}_h}$ and $(w_{k,e}^\mathcal{B})^{\pmb{S}_f}$ in Equations ~\ref{eqn:wcs} and ~\ref{eqn:wfg}, respectively. We don't use these coefficients in our implementation to avoid adding any additional parameter over conventional LSTMs.

In \cite{sperber2017neural}, marginal weights were integrated in the attention mechanism by \textit{biasing attention weights (BATT)} towards lattices nodes with higher marginal weights. Specifically, at decoder step $t$, the attention weight $\alpha_{et}$ corresponding to node $e$ is computed as:
\begin{equation}
  \alpha_{et} = softmax_e(score(\pmb{h}_e, \pmb{s}_t) + \log{w_e^\mathcal{M}})
\end{equation}
where $score(\pmb{.})$ is the alignment model and $\pmb{s}_t$ is the decoder state at decoding step $t$.

Instead of modifying the usual attention mechanism, we propose an alternative approach of scaling the encoder outputs in proportion to marginal weights, which we refer as \textit{weighted encoder output (WEO)}. We omit the biasing in the attention weights computation and update the hidden state output as follows:
\begin{equation}
 \pmb{h}'_e = w_k^\mathcal{M}\cdot \pmb{h}_e
 \label{eqn:weo}
\end{equation}
Note that the modified hidden states in Equation~\ref{eqn:weo} are used only for attention computation and not in the summation of hidden states of predecessor nodes in Equation~\ref{eqn:hsum}.

\subsection{Rescoring Model Architecture}
\label{ssec:architecture}

\begin{figure}
  \centering
  \includegraphics[width=8.6cm]{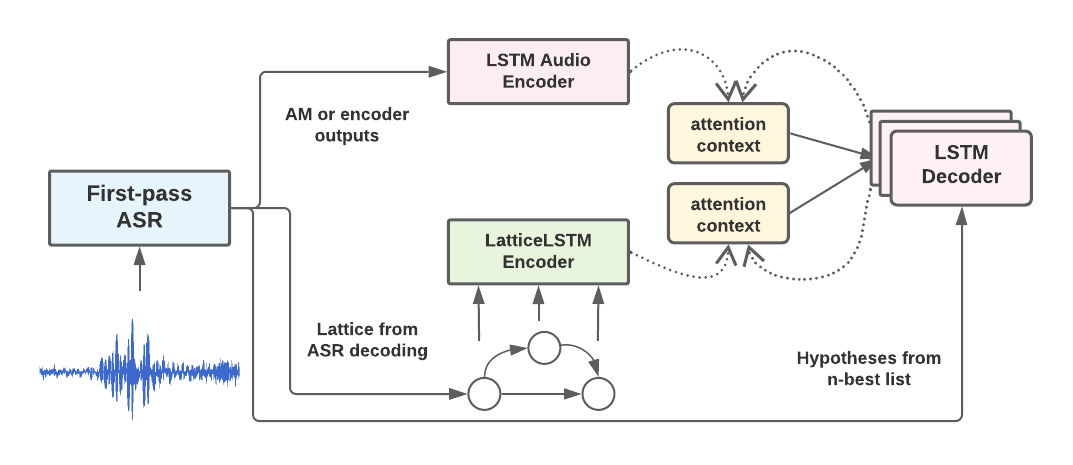}
  \caption{Rescoring model architecture with attention to audio and lattice encoders.}
  \label{fig:architecture}
\vskip -0.3cm
\end{figure}

We use attention-encoder-decoder architecture \cite{bahdanau2015neural} for our rescoring models and experiment with different encoder inputs: 1-best, n-best, audio and lattices. For audio input, we pass acoustic features extracted from the first-pass model to the audio encoder of the rescoring model. In case of n-best input, we encode each hypothesis separately using the same encoder and concatenate their outputs like \cite{hu2020deliberation}. For audio, 1-best and n-best inputs, we use a uni-directional LSTM encoder and for lattice input, we use the (uni-directional) LatticeLSTM encoder from Section~\ref{sec:lattice_encoder}. We also experiment with attention to multiple encoders. If there are more than one encoder in the model, we concatenate the context vectors generated from attention to different encoders. Figure \ref{fig:architecture} shows the rescoring model architecture for attention to both audio and lattice encoders.

\section{Experimental Setup}
\label{sec:setup}

\subsection{Datasets and Architecture Details}
\label{section:datasets}
We used de-identified internal speech data from voice controlled far-field devices for our experiments. We experimented with two different first-pass models: a HMM-based hybrid ASR model and an end-to-end RNN-T model. The hybrid ASR system consits of a LSTM-based acoustic model trained with CTC loss criterion \cite{graves2006connectionist} and a 4-gram LM smoothed with Kneser-Ney \cite{Kneser1995}. Unless specified, hybrid ASR model should be assumed as the first-pass model. Lattices are generated from beam decoding \cite{GenLatticesWFST} of the first-pass models. Apart from the original lattices (referred as \textit{full-lattices}) produced by beam decoding, we also considered pruned \textit{2-best} and \textit{5-best} lattices \cite{bworld}.

The rescoring models contain two LSTM layers with 256 units in the decoder and one LSTM (or LatticeLSTM) layer with 256 units for 1-best/n-best/lattice encoders. For audio encoder, audio features extracted from first-pass ASR are passed to a LSTM layer of 768 units. Both encoder and decoder use an embedding layer with 256 units and vocabulary size of 50k. We use multi-head attention with 4 heads and in case of dual encoders, we use 2 attention heads for each encoder. We also trained a baseline LSTM-based LM with same set of parameters as the decoder of the attention-based rescoring models. The training data used for rescoring models is roughly a quarter of the data used to train first-pass models.

\subsection{Training and Evaluation}
\label{section:training_and_eval}

All the rescoring models, including LSTM-LM, were first trained with MLE loss and then finetuned with MWER loss. Recent works on attention-based rescoring models \cite{sainath2019two,gandhe2020audio} have shown that an additional training with MWER criterion on top of maximum-likelihood training improves accuracy. For evaluation, we run the decoder in the teacher-forcing mode and apply log-linear interpolation of scores of first-pass and rescoring models. The interpolation weight is tuned on a development set. Top 5 hypotheses from the first-pass models are used for n-best rescoring. We evaluate all the experiments on an internal test set containing about 140 hours of audio. We report the results in terms of relative WERR with respect to WER of the first-pass model. Both hybrid and RNN-T baseline first-pass models have absolute WER below 10\%.

\section{Results}
\label{sec:results}
\subsection{Weighting mechanisms in LatticeLSTM}
\label{sec:weighting}
\begin{figure}
     \centering
     \includegraphics[width=0.37\textwidth]{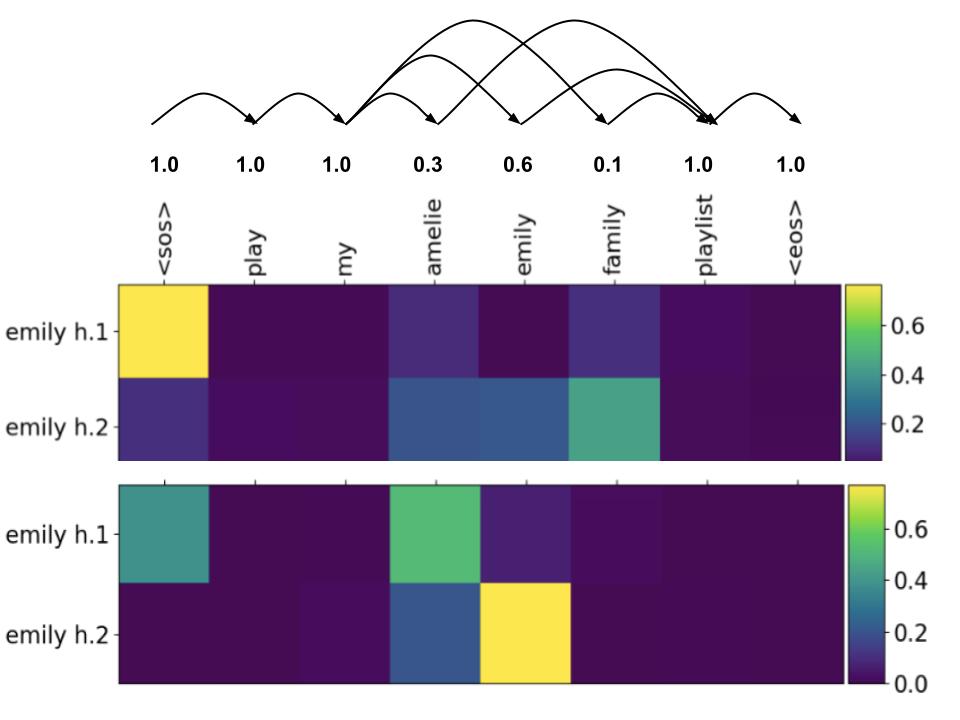}
     \caption{This figure shows attention weights for TreeLSTM (on the top) and weighted LatticeLSTM (on the bottom) models for an example lattice with ground truth as \textit{``play my emily playlist''}. The attention weights correspond to the decoder token \textit{``emily''} in the 1-best hypothesis for two different attention heads in the y-axis. The x-axis shows the linearized lattice nodes fed to the encoder and marginal weights for the nodes are shown on top.}
     \label{fig:attention_weights}
\end{figure}

Table~\ref{tab:weighting_mechanism} shows the effect of the four weighting mechanisms of LatticeLSTM introduced in Section \ref{ssec:weighted_lattice_lstm}. We used \textit{full-lattice} as encoder input for these experiments.
A large degradation is observed with unweighted TreeLSTM lattice encoder for n-best rescoring task. We attribute this to absence of any biasing of the most probable paths in the lattice, causing confusion in the attention module.
This can be seen for an example in Figure~\ref{fig:attention_weights}, where attention weights for TreeLSTM and LatticeLSTM (WCS + BFG + WEO) models are shown for an example lattice corresponding to a confusing token in the decoder. For TreeLSTM model, attention weights are almost uniformly distributed across the homophone variants for one of the attention heads with slightly higher score for the token \textit{``family''}, a common token in the training data. Whereas, for LatticeLSTM model, tokens associated with larger marginal and backward-normalized weights tend to have higher attention weights even if the tokens are rare.
\begin{table}[t]
  \begin{center}
    \caption{Relative WERR over first-pass hybrid ASR model after n-best rescoring using lattice-attention models incorporating different weighting mechanisms in the lattice encoder. The models are trained and evaluated on \textit{full-lattice} input.}
    \label{tab:weighting_mechanism}
    \begin{tabular}{|c|c|}
      \hline
      \textbf{Weighting Mechanism} & \textbf{WERR (\%)}\\
      \hline
      None (TreeLSTM)  &  -8.4\\
      \hline
      WCS & 2.4\\
      BFG  & 3.8\\
      BATT & 5.0\\
      WEO & 4.9\\
      \hline
      WCS + BFG + BATT & 5.0\\
      WCS + BFG + WEO & 5.0\\
      \hline
    \end{tabular}
  \end{center}
  \vskip -0.1cm
\end{table}

Introducing any of the weighting mechanisms significantly reduces the degradation observed with TreeLSTM for n-best rescoring. Weighting mechanisms based on marginal weights (i.e. WEO and BATT), have a larger impact than the mechanisms incorporating backward-normalized weights. From here on, lattice encoders should be assumed to have WCS, BFG and WEO weighting mechanisms.

\subsection{Impact of lattice depth}
\label{ssec:weighting}

Table~\ref{tab:lattice_depth_table} shows relative WERR over first-pass WER for rescoring models trained on lattices with varying depths. Note that for a \textit{n-best} lattice, the original lattice is pruned to top \textit{n} hypotheses in both training and evaluation. \textit{2-best} lattice as the encoder input results in 4.6\% WERR and \textit{5-best} or \textit{full} lattice inputs provide very small further improvements. This is due to most lattices being shallow as only 31\% of the lattice have more than two alternate hypotheses in our data. In Table~\ref{tab:lattice_depth_table}, we also report results on partitions of the test set with $\leq$2 and \textgreater2 alternate hypotheses in the original lattice. All three models have similar performance on utterances with $\leq$2 alternatives but on utterances with \textgreater2 alternate hypotheses, models with lattices of higher depths as encoder input perform better. This suggests that models leveraging more alternative hypotheses from the lattice can benefit from the richness of the lattice.

\begin{table}[t]
  \begin{center}
    \caption{Relative WERR (\%) over first-pass hybrid ASR model after n-best rescoring using lattice-attention models trained on lattices of different depths. We report WERR on full test set and utterances with $\leq$2 and \textgreater2 alternate hypotheses in the original lattices.}
    \label{tab:lattice_depth_table}
    \begin{tabular}{|c|c|c|c|}
      \hline
      \textbf{Lattice Encoder} & \textbf{Full} & \textbf{Utts with} & \textbf{Utts with}\\
       & \textbf{test set} & \textbf{$\leq$2 hyps} & \textbf{$>$2 hyps}\\
      \hline
      2-best lattice & 4.6 & 3.8 & 4.8\\
      5-best lattice & 4.8 & 3.8 & 5.1\\
      Full-lattice & 5.0 & 3.8 & 5.5\\
      \hline
    \end{tabular}
  \end{center}
\vskip -0.5cm
\end{table}

\subsection{Encoder type}
\label{ssec:encoders}
\begin{table}[t]
  \begin{center}
    \caption{Relative WERR (\%) over first-pass models (Hybrid and RNN-T) after n-best rescoring using different models.}
    \label{tab:encoder_type_table}
    \begin{tabular}{|c|c|c|}
      \hline
      \textbf{Rescoring Model} &  \textbf{Hybrid} & \textbf{RNN-T}\\
      \hline
      No encoder (LSTM-LM) & 2.9 & 1.2\\
      \hline
      1-best encoder & 3.8 & 2.7\\
      5-best deliberation encoder & 4.1 & 3.1\\
      \hline
      5-best lattice encoder & 4.8 & 3.6\\
      Full-lattice encoder & 5.0 & 3.8\\
      \hline
      Audio encoder (LAS) & 5.2 & 3.7\\
      Audio \& 1-best encoders & 6.5 & 4.4\\
      Audio \& 5-best deliberation encoders & 6.9 & 4.8\\
      \hline
      Audio \& 5-best lattice encoders & 7.6 & 5.5\\
      Audio \& Full-lattice encoders & 7.8 & 5.7\\
      \hline
      Oracle at 5-best & 42.9 & 36.1\\
      \hline
    \end{tabular}
  \end{center}
\vskip -0.5cm
\end{table}

Table~\ref{tab:encoder_type_table} captures relative WERR of rescoring models with different encoder types over first-pass WER. Except LSTM-LM, all the models are attention-based and trained spearately on audio features/1-best/n-best/lattice outputs of hybrid and RNN-T first-pass models. The LSTM-LM rescoring model is same for both hybrid and RNN-T as it is trained only on transcriptions. Our results show that attention to any of the encoder provides more improvement compared to discriminatively trained LSTM-LM. The consistently smaller improvement for RNN-T system compared to hybrid is due to stronger first-pass model, evident from 1.2\% WERR for RNN-T compared to 2.9\% WERR for hybrid when rescored with same LSTM-LM model. The lattice encoder performs better compared to 1-best or n-best encoders and provides similar WERR compared to audio encoder, agnostic of the first-pass model. In situations where the audio is not available for second-pass rescoring, lattice encoders could be a good proxy. The audio-attention model can be seen as LAS rescoring proposed in \cite{sainath2019two} and audio \& n-best attention can be seen as deliberation network rescoring of \cite{hu2020deliberation} without the joint training of first-pass and second-pass models. The LatticeLSTM-based \textit{5-best} lattice encoder outperforms deliberation-style \textit{5-best} encoder of \cite{hu2020deliberation} for both single and dual encoder setups. Also, due to compactness of lattice representation, there are 52\% fewer forward-passes on average in LatticeLSTM encoder compated to deliberation network. Adding attention to additional 1-best, n-best or lattice encoder provides further WERR over audio-only attention model. The model with attention to both, audio and \textit{full} lattice, achieves the best result with 7.8\% relative WERR over first-pass for hybrid and 5.7\% for RNN-T. The small difference between \textit{5-best} or \textit{full} lattice encoders can be attributed to very small number of lattices with more than five alternatives as discussed in Section \ref{ssec:weighting}.
\section{Conclusions}
\label{sec:conclusions}

We proposed an attention encoder-decoder model with attention to lattices for rescoring n-best hypotheses generated by an ASR model. The lattice-attention rescoring model achieves 4-5\% relative WERR over hybrid or RNN-T first-pass models, which is comparable to performance of audio-only attention model. Attention to both, audio and lattices, brings further improvement, resulting in 6-8\% WERR. We showed that LatticeLSTM-based lattice encoder excels over n-best encoder respresentation of deliberation network.  Further, deeper lattices can benefit from richness of the lattice. As opposed to the attention biasing method, we proposed a simpler alternative with similar performance, in which encoder outputs are scaled in proportion to lattice weights while keeping the usual attention mechanism unchanged. We also looked into different weighting mechanisms in the lattice encoder and showed that incorporating weights in the lattice encoder is essentital for attention-based models to avoid inherent confusions in the lattices arising from conflicting arcs.

%\vfill
%\pagebreak

\bibliographystyle{IEEEbib}
\bibliography{strings}

\end{document}